\documentclass{article}

\usepackage{graphicx}
\usepackage{tabularx}
\graphicspath{{figure/}}
\usepackage{amsmath}
\usepackage{xcolor}
\usepackage{subcaption}
\usepackage[bookmarks=true]{hyperref}
\usepackage{float}
\usepackage{siunitx} 
\usepackage{comment}
\usepackage{authblk}

\newcommand{\figref}[1]{Figure~\ref{#1}}
\newcommand{\tabref}[1]{Table~\ref{#1}}

\newcommand{\fname}[1]{{\texttt{fairDetect}}}
\newcommand{\insitu}{{\emph{in situ}}}
\newcommand{\Insitu}{{\emph{In situ}}}

\newcommand{\ome}{{$\omega$}}
\newcommand{\xl}{{$x_L$}}
\newcommand{\yl}{{$y_L$}}

\usepackage{siunitx} 
\newcommand{\degs}[1]{{{#1}$^\circ$}}

\title{Rapid detection of rare events from \emph{in situ} X-ray diffraction data using machine learning}

\author
{Weijian Zheng,$^{\dagger}$ Jun-Sang Park,$^{\dagger}$ Peter Kenesei, Ahsan Ali, Zhengchun Liu, Ian T. Foster, Nicholas Schwarz,  Rajkumar Kettimuthu, Antonino Miceli, Hemant Sharma $^{\ast}$\\
\small{Argonne National Laboratory, 9700 S Cass Ave, Lemont, IL, USA}\\
\small{$^\ast$To whom correspondence should be addressed: hsharma@anl.gov}\\
\small{$^\dagger$Equal contributions}
}

\date{\today}

\begin{document}

\maketitle

\begin{abstract}
High-energy X-ray diffraction methods can non-destructively map the 3D microstructure and associated attributes of metallic polycrystalline engineering materials in their bulk form. These methods are often combined with external stimuli such as thermo-mechanical loading to take snapshots over time of the evolving microstructure and attributes. However, the extreme data volumes and the high costs of traditional data acquisition and reduction approaches pose a barrier to quickly extracting actionable insights and improving the temporal resolution of these snapshots. Here we present a fully automated technique capable of rapidly detecting the onset of plasticity in high-energy X-ray microscopy data. Our technique is computationally faster by at least 50 times than the traditional approaches and works for data sets that are up to 9 times sparser than a full data set. This new technique leverages self-supervised image representation learning and clustering to transform massive data into compact, semantic-rich representations of visually salient characteristics (e.g., peak shapes). These characteristics can be a rapid indicator of anomalous events such as changes in diffraction peak shapes. We anticipate that this technique will provide just-in-time actionable information to drive smarter experiments that effectively deploy multi-modal X-ray diffraction methods that span many decades of length scales.
\end{abstract}

\section{Introduction}
Metals play a significant role in modern society. They are used in a wide range of applications such as transportation; construction; energy generation, storage and delivery; and security. The performance requirements for these polycrystalline engineering materials have pushed the diversity and complexity of processes and elements employed in those materials to optimize structure and properties \cite{Greenfield2013, Graedel2015}. Frameworks like integrated computational materials engineering \cite{Olson2000} and materials informatics \cite{Rajan2005} aim to accelerate material discovery and process optimization so that new materials that meet performance requirements can be deployed more quickly. Calibration and validation of these frameworks rely on microstructure and associated attributes acquired across multiple length scales at different conditions.

High-energy synchrotron X-ray ($>=$50 keV) diffraction methods are capable of non-destructively characterizing metallic polycrystalline materials in their bulk form. In particular, high-energy diffraction microscopy (HEDM) can extract 3D microstructure information and grain-resolved attributes; it can also track their evolution when combined with \insitu{} thermo-mechanical loading capabilities \cite{Lienert.2011up, Schuren.2015dr, naragani.2017rp}. The far-field HEDM (FF-HEDM) variant \cite{Bernier2020, Park2021} can provide the center of mass, crystallographic orientation, and elastic strain tensor for each constituent grain in a polycrystalline aggregate. These FF-HEDM measurements are often combined with other measurement modalities such as near-field HEDM and tomography to obtain a fuller picture of the microstructure and state and their evolution in a polycrystalline material \cite{Suter.2008kn, Turner.2016xu, naragani.2017rp, sangid.2020wm}.

In a typical \insitu{} HEDM experiment, the macroscopic stimulus to be applied on a sample is decided \emph{a priori}, based for example on a known response relationship. For instance, in an \insitu{} HEDM experiment where a sample is subject to various levels of mechanical loading to study the material response heterogeneity at the mesoscale, the levels -- at which the loading is paused to deploy a higher resolution and more beamtime-consuming characterization method -- are often decided based on macroscopic milestones such as yield strength in the macroscopic stress-strain curve of the material. Alternatively, such experimental decisions to be made on-the-fly may require heroic efforts and significant human involved processing during time-limited beam access to reduce the full HEDM data set and identify when the scientifically interesting phenomena occur \cite{naragani.2017rp, Ravi2021, Suter.2008kn, Li2023, maddali.2020bc, simmons.2015te}. Hence, the high-level of reliable automation is desirable and crucial for a successful study.

\begin{figure}[H]
    \centering
    \includegraphics[width=1\textwidth]{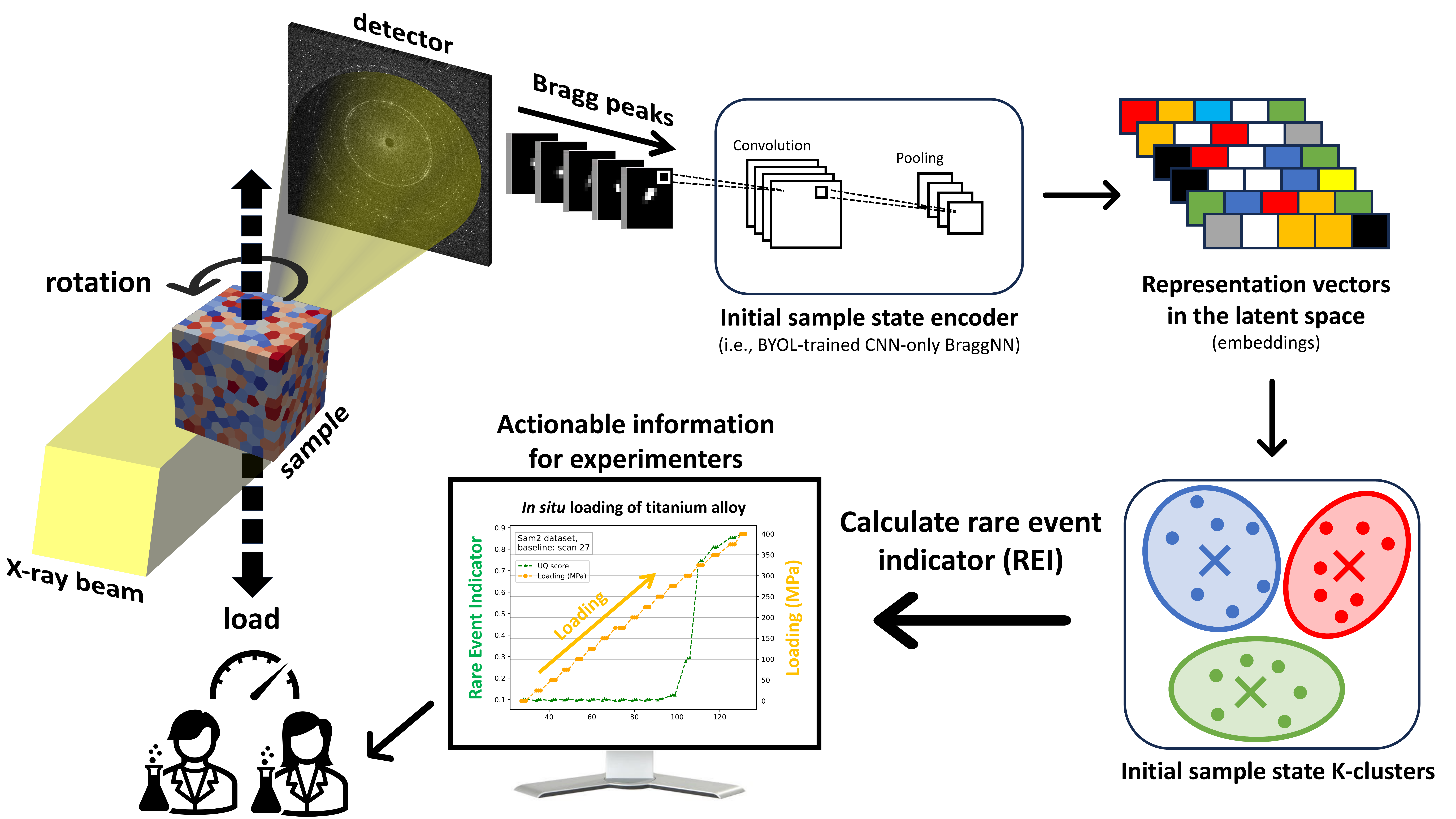}
    \caption{An illustration of the workflow for rapid inference of microstructural deformation. A polycrystalline sample is subject to mechanical loading while we acquire HEDM data. Bragg diffraction spots from the initial material state are used to train an image representation model (encoder) and a clustering model. These two models combined are sensitive to changes in the diffraction spots. Using these models, the rare event indicators are computed as we continue to apply mechanical loading to the sample and acquire HEDM data. A significant increase in REI is quantitative actionable information that the experimenters can use to steer the course of the experiment.}
    \label{fig:overall-flow}
\end{figure}

Here we propose a robust, self-supervised, machine learning-based framework (\figref{fig:overall-flow}) that enables rapid identification of, and thus automated response to, the minute changes in diffraction spots measured by FF-HEDM and probable microstructural changes in polycrystalline metals. This framework is particularly timely with the improved brilliance and coherence of the next generation diffraction limited synchrotron X-ray sources and significant advances in detector technologies capable of rapid and accurate measurement of single photon events --- developments that allow many mesoscale measurement techniques like HEDM to be combined with nanoscale techniques like Bragg coherent diffraction imaging (BCDI) and dark-field X-ray microscopy (DFXM) in a single instrument \cite{maddali.2020bc, aps_u_fdr_report_www}. Our framework can provide quantitative actionable information about material state and guide experimenters as to when to deploy these higher resolution techniques. We refer to this actionable information as rare event indicator (REI). Using a variety of \insitu{} FF-HEDM data, we demonstrate that the proposed framework and REI is capable of rapidly detecting mechanical yielding, an eventual but rare event that experimenters often seek to capture through \insitu{} FF-HEDM experiments. 

\section{Results and discussion}
\label{sec:results_and_discussion}
We acquired four sets of \insitu{} FF-HEDM diffraction patterns using an area detector. \S\ref{subsec:hedm} provides the salient features of these HEDM experiments. The first set of patterns was acquired with a 304L-stainless steel (304L-SS) sample with a face-centered cubic crystal symmetry. The second set of patterns was acquired with a Ti-6V-4Al (Ti64) sample with a hexagonal close-packed crystal symmetry. These two data sets were acquired in uniaxial tension. The third data set used a pack of sand with a trigonal crystal symmetry under compression. The fourth data set was acquired on a Ti-7Al (Ti7Al) alloy sample with a hexagonal close-packed crystal symmetry under continuous uniaxial tension.

We anticipated the 304L-SS, Ti64, and Ti7Al samples to accommodate mechanical yielding or plastic deformation by crystallographic slip \cite{taylor.1938pu} resulting in diffraction spot smearing \cite{Obstalecki.2014pk} while the sand sample to exhibit brittle fracture with minimal diffraction spot smearing. For the sand sample, we expected the number of diffraction spots to increase with the applied load as the sand particles fracture into smaller but still coherent pieces, therefore the shape of resulting diffraction spots was expected to exhibit minimal or no smearing at all, while their intensity would show a dramatic decrease. With these expectations, our rare event detection framework, which consists of an image representation model and a clustering model, was trained using various permutations of diffraction patterns acquired at zero load -- zero strain state (reference state) from each sample. These framework permutations were deployed to assess whether yielding could be detected reliably. Furthermore, the sensitivity of our framework and REI to instrumental changes was assessed using the 304L-SS FF-HEDM patterns acquired using a variety of beam sizes and incident X-ray photon flux levels.

\subsection{\Insitu{} FF-HEDM of stainless steel}
\label{subsec:ss_uq}
\begin{figure}[tbhp]
    \centering
    \includegraphics[width=0.8\textwidth]{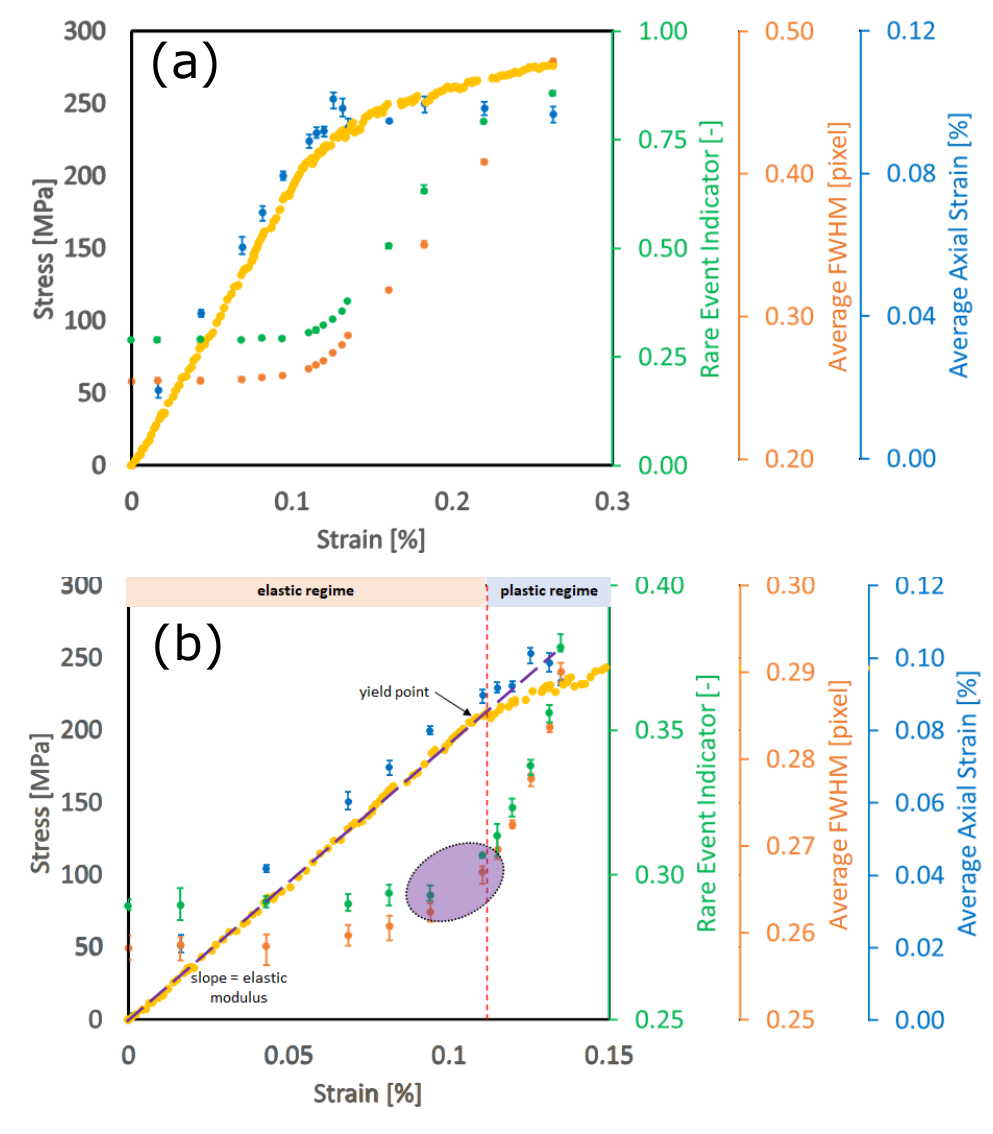}
    \caption{Applied stress, REI, average FWHM, and average axial lattice strain vs. strain for (a) the entire \insitu{} FF-HEDM experiment (b) magnified view near the elastic-plastic regime. The error bars for REI, FWHM, and lattice strain at a target stress level indicate the range of respective metrics observed in the four \SI{0.4}{\milli\meter}-tall material volumes. The dotted yellow curve shows the stress-strain curve of the 304L-SS sample measured during \insitu{} FF-HEDM experiment.}
    \label{fig:steel_rei}
\end{figure}

The dotted yellow curve in \figref{fig:steel_rei} shows the typical stimulus (stress) -- response (strain) relationship of a metallic material. At lower stresses, the relationship between macroscopic stress and strain is nominally linear and elastic, but at larger stresses, this is no longer the case. The level of stress where the linear relationship between stress and strain breaks from linearity is often referred to as the macroscopic yield point or yield stress of the material (approximately 225 MPa in this case). Here, the stress--strain curve was acquired during our \insitu{} FF-HEDM experiment under uniaxial tension. The geometry of the 304L-SS sample was identical to that described in \cite{Wang2022} with \SI{11}{\milli\meter} (length) $\times$ \SI{1}{\milli\meter} (width) $\times$ \SI{1}{\milli\meter} (thickness) in the gauge section. Gold markers were attached along the gauge section \cite{shade.2016pf} so that identical material volume could be measured throughout the experiment. 

In this \insitu{} FF-HEDM experiment, there were four target stress levels in the elastic regime, five in the elastic-plastic transition regime (the so-called ``knee'' of the stress-strain curve), and four in the plastic regime well beyond the knee. Each target stress level was reached through a constant displacement rate of \SI{1.1}{\micro\meter/\s} to achieve a nominal strain rate of \SI{0.0001}{\s^{-1}}; once the target stress level was reached, the applied stress was intentionally relieved by approximately 10\% before carrying out the FF-HEDM pattern acquisition to prevent further changes to the material by stress relaxation during the scans\footnote{The sample was under displacement control during FF-HEDM pattern acquisition; applied stress level fluctuation was minimal during a FF-HEDM scan even in the plastic regime.}. 

An x-ray beam size of \SI{2}{\milli\meter} $\times$ \SI{0.4}{\milli\meter} was used for FF-HEDM pattern acquisition. The width of the x-ray beam ensured that a \SI{0.4}{\milli\meter}-tall material volume in the sample gauge section was always illuminated by x-rays during an FF-HEDM pattern acquisition. At each target load level, four sets of FF-HEDM pattern acquisition were executed illuminating a total of \SI{1.6}{\milli\meter} tall contiguous volume (4 layers) in the sample gauge section. The gold markers in the gauge section ensured that the identical material volume was interrogated across multiple target load levels reached during \insitu{} loading. 

At each target load level, FF-HEDM patterns were used to compute the REI\footnote{Here, the FF-HEDM patterns acquired at the zero load - zero strain point were used as the baseline data set to train the image representation model. For more details on the image representation model, the readers are referred to \S\ref{subsec:event_detection}.}, average diffraction spot full-width at half maximum (FWHM) in the azimuthal direction, and average axial lattice strain\footnote{The FWHM and axial lattice strain are the more commonly used metrics to detect the rare event, namely yielding in this case.}; \figref{fig:steel_rei} shows the evolution of these metrics with applied stress denoted by the green dots, the orange dots, and the blue dots, respectively. Here, the image representation model and the clustering model to compute the REI were trained using the 304L-SS sample's reference state diffraction pattern. The FWHM and axial lattice strain were computed using MIDAS \cite{Sharma2012ASetup,Sharma2012AGrains}. All three metrics, namely REI, average diffraction spot FWHM in the azimuthal direction, and average lattice strain require the extraction of diffraction peaks from 2D diffraction patterns, with the subsequent steps being inference (for REI, requiring $\sim$2s computation time), peak-shape fitting (for FWHM, requiring $\sim$100s computation time) and full microstructure reconstruction (for lattice strain, requiring $\sim$500s computation time).

We demonstrate in this paper that REI can provide actionable information comparable to the traditional metrics for polycrystalline metallic alloys subjected to mechanical loading. \figref{fig:steel_rei} (b) is a magnified view of \figref{fig:steel_rei} (a) near the elastic and elastic-plastic transition regime. In this regime, researchers often need to make the difficult decision to pause the loading and conduct higher resolution measurements and allocate precious beam time based on limited \emph{a priori} knowledge about the material (such as stress-strain curve obtained \emph{ex situ}) or relying on FWHM or lattice strain metrics that require FF-HEDM patterns acquired over a larger angular range, substantial computing resources, and time. In general, the three metrics follow the trends observed in the macroscopic stress-strain curve, with significant inflections occurring near the knee of the stress-strain curve. REI and FWHM remain fairly constant in the elastic regime. As the material approaches its macroscopic yield point, REI and FWHM both start to increase, most notably in the region highlighted by the purple ellipse, roughly coinciding with the macroscopic yield point. They continue to increase substantially as the sample transitions into the plastic regime. On the other hand, the average axial lattice strain increases linearly in the elastic regime. It breaks from linearity near the macroscopic yield point and then remains relatively constant in the plastic regime\footnote{This is expected as lattice strain is measuring only the elastic strain (change in the lattice spacing).}. This figure illustrates that the REI is a metric that is sensitive to material yielding accommodated by crystallographic slip and delivers information comparable to the traditional metrics while being at least 50 times faster to compute.

\subsection{\Insitu{} FF-HEDM of titanium alloy}
\label{subsec:ti_rei}
\begin{figure}[htbp]
    \centering
    \includegraphics[width=0.9\textwidth]{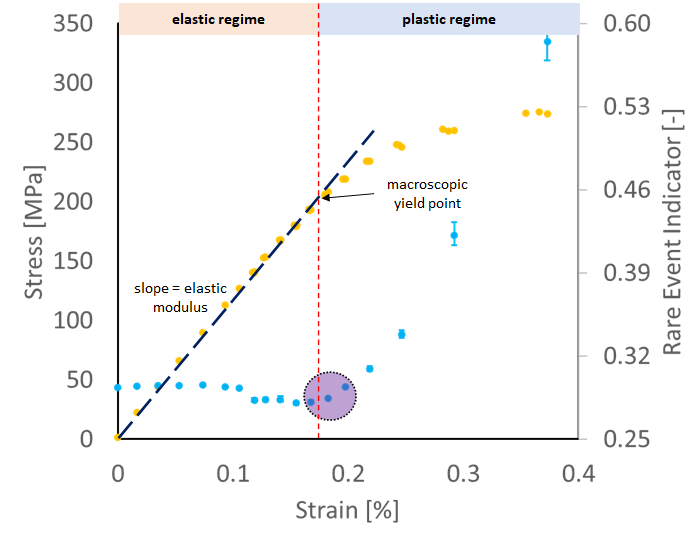}
    \centering
    \caption{The \insitu{} stress-strain curve of the Ti64 alloy sample (yellow dots) and associated REI results (blue dots) in the elastic and elastic-plastic transition regimes. REI remains roughly constant in the elastic regime. As the material approaches its macroscopic yield point, REI also starts to increase significantly highlighted by the purple ellipse, and continues to increase as the plastic regime sets in. The abrupt change in REI near 0.1\% strain is due to a beam size change between two target load levels; this is discussed further in \S\ref{subsec:rei_robustness_study}.}
    \label{fig:ti_s_rei_elastic_full} 
\end{figure}

\figref{fig:ti_s_rei_elastic_full} shows the stress-strain curve and rare event indicator results for a Ti-6V-4Al (Ti64) alloy, magnified in the elastic and elastic-plastic transition regime. The dotted yellow curve shows the stress-strain curve of the Ti64 material in uniaxial tension acquired during \insitu{} FF-HEDM experiment; it shows similar characteristics as the stress-strain curve for the 304L-SS sample (\figref{fig:steel_rei}) with the elastic-plastic transition occurring approximately at 200 MPa. Here, the \insitu{} FF-HEDM experiment was conducted using the Rotation and linear Axial Motion System (RAMS) load frame \cite{Schuren.2015dr}. The sample geometry was identical to the one presented in \cite{Menasche2021} with a \SI{1}{\milli\meter} $\times$ \SI{1}{\milli\meter} cross section in the gauge section. Gold markers were attached in the gauge section \cite{shade.2016pf} to keep the illuminated material volume consistent throughout the \insitu{} experiment. The experimental procedure was similar to that used for the 304L-SS sample; the nominal strain rate was \SI{0.0001}{\sec^{-1}} and loading was temporarily paused to collect FF-HEDM patterns along multiple layers in the sample gauge section. 

The blue dots in \figref{fig:ti_s_rei_elastic_full} show the evolution of REI when the image representation model and clustering model were trained by using the 304L-SS FF-HEDM patterns acquired at zero load-zero strain (reference state). The overall evolution of REI is consistent with that observed in the 304L-SS sample; REI remains relatively constant in the elastic regime and then starts to increase near the knee of the stress-strain curve demonstrating that the image representation model and the clustering model (and therefore REI) is sensitive to materials accommodating plastic deformation through crystallographic slip. Furthermore, \figref{fig:ti_s_rei_elastic_full} illustrates that those models are transferable between materials that share a similar deformation mode. The abrupt decrease in REI  at approximately 0.1\% strain (or 150 MPa applied stress) is due to a change made at this point in the beam size used for FF-HEDM experiments in order to alter the vertical spacing between layers. The robustness and sensitivity of REI are examined further in \S\ref{subsec:rei_robustness_study}.

\begin{figure}[htbp]
    \centering
    \includegraphics[width=0.75\linewidth]{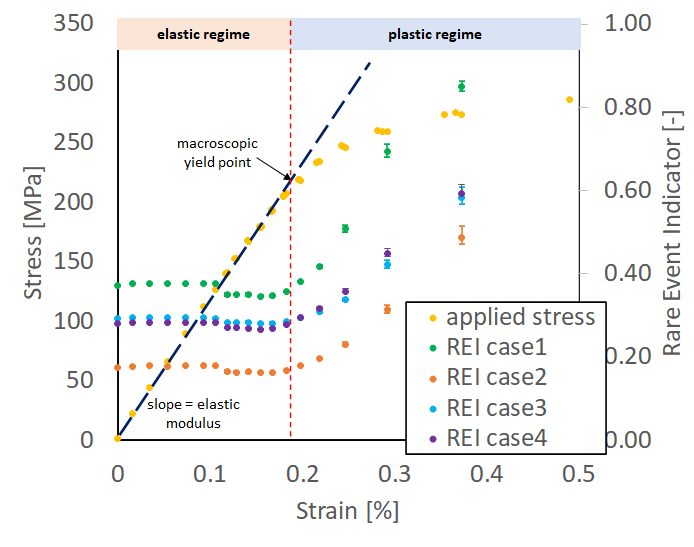}
    \caption{The \insitu{} stress-strain curve and REI evolution for the Ti64 sample in the elastic and elastic-plastic transition regime.  Four cases of REI using permutations of training reference state patterns (\tabref{tab:Ti64_rei_cases}) are presented.}
    \label{fig:ti_svse_rei} 
\end{figure}

We computed several cases of REI in order to examine the performance of the rare event detection framework and the choice of baseline and reference data (\figref{fig:ti_svse_rei}). \tabref{tab:Ti64_rei_cases} summarizes the permutations of training reference state patterns to compute the REI cases. In all four cases, we used the optimized hyperparameter values determined from the 304L-SS experiment. Naturally, the absolute value associated with each REI case is different between the four cases as the training data sets are different, however, the four cases all share a similar trend in that they remain relatively constant in the elastic regime, start to increase near the elastic-plastic regime, and continues to increase in the plastic regime. The inflection point where the REI values start to increase is consistently near the macroscopic yield point. This highlights that the image representation and clustering models trained on a data set that shares a similar deformation mode but collected on a different sample or material are transferable. While it is recommended that the patterns acquired before the beginning of \insitu{} loading be used as the training set, such transferability can be useful when using REI to estimate the material state, for instance in a high-throughput measurement setup.

\begin{center}
\begin{table}
    \begin{tabular}{ c || p{4.5cm} | p{4.5cm} }
    \hline
    REI Case & Training reference state patterns for encoder & Training reference state patterns for clustering model \\ 
    \hline
    \hline
    1 & Ti64 sample & Ti64 sample \\
    2 & Ti64 sample & 304L-SS sample \\
    3 & 304L-SS sample & 304L-SS sample \\
    4 & 304L-SS sample & Ti64 sample \\
    \hline
    \end{tabular}
    \caption{The reference state patterns employed to train the encoder and clustering models for the REI cases presented in \figref{fig:ti_svse_rei}.}
    \label{tab:Ti64_rei_cases}
\end{table}
\end{center}

\subsection{\Insitu{} FF-HEDM of sand}
\label{subsec:sand_uq}
In \S\ref{subsec:ti_rei}, we established that the image representation and clustering models trained using the 304L-SS FF-HEDM patterns acquired at reference state can be used to detect rare events and anomalies in other material systems that share a similar deformation modality. Here, we introduce a set of \insitu{} diffraction patterns acquired from a material system that does not accommodate deformation by crystallographic slip. 

\begin{figure}[htbp]
    \centering
    \includegraphics[width=0.9\textwidth]{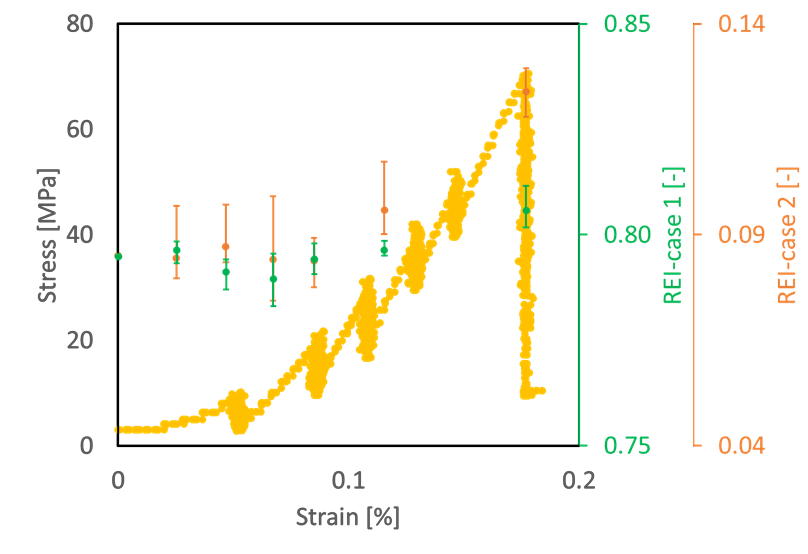}
    \caption{Stress-strain curve and REI evolution for the sand sample. Stress and strain are compressive mimicking the compaction that sand is typically subjected to. The stress-strain curve does not exhibit the characteristics illustrated in \figref{fig:steel_rei} and \figref{fig:ti_svse_rei}. The orange dots show the REI case where the sand reference state patterns were used as the training data set. The green dots show the REI case where the 304L-SS reference state patterns were used as the training data set. Both REI cases show minimal changes as the sand sample is subject to compressive loads up to catastrophic failure of the material system.}
    \label{fig:sand_s_uq_vs_s} 
\end{figure}

\figref{fig:sand_s_uq_vs_s} shows the stress-strain curve and associated REI for an aggregate of sand subject to compression. The dotted yellow curve shows the stress-strain curve of sand aggregate in unaxial compression. The \insitu{} mechanical loading procedure was similar to the one described in \cite{hurley.2017ys}. The sample fractured catastrophically when compressed beyond 75 MPa. The sample also does not exhibit the elastic, elasti-plastic transition, and plastic regimes that were observed in the 304L-SS (\figref{fig:steel_rei}) and Ti64 (\figref{fig:ti_s_rei_elastic_full}) samples. Instead, the stress-strain curve shows a constant increase and an abrupt decrease in applied stress indicative of catastrophic failure. We consider two REI cases similar to \figref{fig:ti_svse_rei}. For REI Case 1 (green dots), the image representation model and the clustering model are both trained using the 304L-SS reference state patterns. For REI Case 2 (orange dots), these models are trained using the sand reference state patterns. In both cases, the hyperparameters were optimized based on the 304L-SS patterns. While the absolute magnitudes are different, the two REI cases show minimal changes with respect to loading. This implies that the underlying architecture in our rare event detection framework is tuned to detect anomalies in diffraction spots induced by crystallographic slip but it is insensitive to deformation carried by fracturing of grains.

\subsection{Robustness to experiment configurations}
\label{subsec:rei_robustness_study}

We have demonstrated that REI is sensitive to changes in plasticity in the materials, but there are additional experiment configurations that can lead to changes in REI too. Here we investigate the sensitivity of REI to different configurations during the 304L-SS experiment in $\S$\ref{subsec:ss_uq}. Figure \ref{fig:steel_rei} includes data collected with the following configuration: 0.4 mm x-ray beam height, rotation angle start at -180$^\circ$, rotation angle step 0.25$^\circ$, this was the reference state for computing the absolute change in REI ($\Delta$REI). \tabref{tab:sensitivity} shows $\Delta$REI as a result of different changes in experiment configurations. For each $\Delta$REI calculation, the material deformation state was constant, only the corresponding experiment configuration was changed.

Changes in the starting rotation angle and rotation angle step change the position and slicing of diffraction spots (3-dimensional) on individual frames (2-dimensional). Since our model works with diffraction signals on individual frames instead of 3-D shapes\footnote{Using 2-D spots instead of 3-D allows for improvements in computation speed and working with streaming data.}, these changes are expected to change REI as well. \tabref{tab:sensitivity} shows that changing the starting rotation angle \footnote{Offsets of $\pm$0.125$^\circ$, $\pm$0.25$^\circ$, $\pm$0.314$^\circ$ and $\pm$0.628$^\circ$ in starting rotation angle were used.} has the smallest influence on REI. We will use this $\Delta$REI as a baseline (denoted $\Delta$REI$_{b}$) to compare against other experimental parameters. Changing the rotation angle step (0.1$^\circ$ vs 0.25$^\circ$) has more than 3 times larger $\Delta$REI as compared to $\Delta$REI$_{b}$.

Changing the position of data acquisition in the sample (4 different positions) results in 2 times larger $\Delta$REI than $\Delta$REI$_{b}$. This is attributable to local changes in the microstructure at any given sample state. On the other hand, changes in other experiment configurations (with different positions in the sample) lead to more than 4 times higher $\Delta$REI compared to $\Delta$REI$_{b}$: 4 times for changing x-ray beam size (0.1, 0.2, and 0.4 mm), 5 - 10 times for using sub-optimal incident x-ray flux for low and high plastic deformation\footnote{Here, optimal incident x-ray flux is the setting in which the full dynamic range of the area detector is utilized without acquiring saturated pixels during a FF-HEDM scan; sub-optimal incident x-ray flux means the dynamic range of the detector is underutilized.}\footnote{High local microstructure changes from deformation also contribute to high $\Delta$REI.} and $\sim$6 times higher for changing both x-ray beam size and flux.

Furthermore, for the titanium data set (Figure \ref{fig:ti_s_rei_elastic_full}) the change in REI near 0.1\% strain by $\sim$1.3$\times$10$^{-2}$ is because of a change in x-ray beam size from 0.4 mm to 0.2 mm.

$\Delta$REI for detecting the onset of plastic deformation in Figure \ref{fig:steel_rei} is 1.38$\times$10$^{-2}$, the difference between REI of two sample states in the purple ellipse. From \tabref{tab:sensitivity}, we can see that only changes in the starting rotation angle and position in the sample yield a $\Delta$REI smaller than this value. This indicates that an experiment utilizing REI as an indicator can vary these two experiment configurations, namely, starting rotation angle and position in the sample, but must keep the other experiment configurations constant during the experiment: rotation angle step, x-ray beam size, and x-ray beam flux.

\begin{center}
    \begin{table}[tbhp]
        \begin{tabular}{ p{6.5cm} || c | c  }
        \hline
         & $\Delta$REI & Average REI  \\ 
        Experiment change & $[ \times 10^{-2}]$ & $[-]$ \\ 
        \hline
        \hline
        Different starting rotation angle & 0.52 & 0.27	\\
        Different rotation angle step & 1.7	& 0.28  \\
        Different position & 1.1 & 0.31 \\
        Different position and x-ray beam size & 2.1 & 0.28 \\
        Different position and x-ray beam flux & 2.7 & 0.29\\
        Different position and x-ray beam flux, large plastic deformation & 5.2	& 0.52 \\
        Different position, x-ray beam size, and beam flux & 2.9	& 0.27 \\
        \hline
        \end{tabular}
        \caption{Sensitivity to experiment parameters}
        \label{tab:sensitivity}
    \end{table}
\end{center}

\subsection{REI using partial data set}
\label{subsubsec:_partial_data_ss_uq}
The ability to monitor the material state changes and provide actionable information in real-time predicates having a sufficient number of frames to extract reliable REI quickly. To estimate the minimum number of detector frames necessary to provide a reliable rare event indicator, we used the FF-HEDM patterns from the 304L-SS sample acquired over \ome-range of \SI{360}{\degree}. For the FF-HEDM patterns acquired at a given target load level, 
\begin{enumerate}
    \item A starting \ome~angle (or a frame in the stack of FF-HEDM patterns acquired over \ome-range of \SI{360}{\degree}) was chosen randomly, 
    \item A contiguous segment of frames corresponding to a particular $\Delta\omega$ range starting from the chosen \ome~angle is extracted from the FF-HEDM patterns, 
    \item REI for the contiguous segment is computed, and 
    \item This procedure is repeated 20 times with random starting \ome~angles to estimate the range of REI values arising from using a partial FF-HEDM pattern set.
\end{enumerate}

\begin{figure}[htbp]
        \centering
        \includegraphics[width=0.75\linewidth]{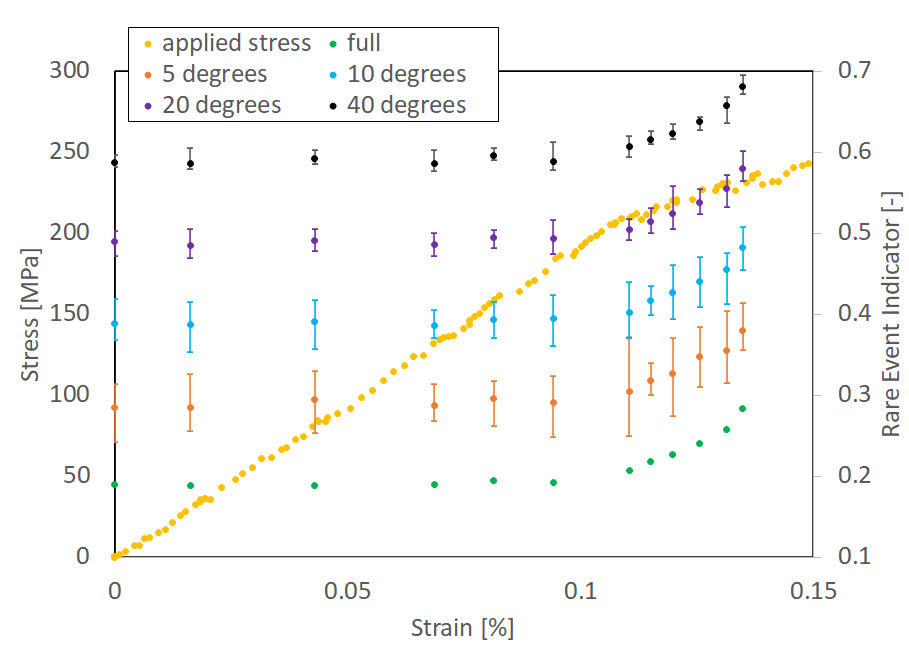}
    \caption{The \insitu{} stress-strain curve and REI evolution. The \insitu{} stress-strain curve and REI evolution were computed using partial data sets for the 304L-SS sample in the elastic and elastic-plastic transition regime. For clarity, the REI values are intentionally shifted by -0.1, 0, 0.1, 0.2, and 0.3 for $\Delta\omega$ of \SI{360}{\degree}, \SI{5}{\degree}, \SI{10}{\degree}, \SI{20}{\degree}, and \SI{40}{\degree}, respectively. }
    \label{fig:partial_dataset_results} 
\end{figure}

\figref{fig:partial_dataset_results} presents a magnified view of the elastic and elastic-plastic transition regime for $\Delta\omega$ of \SI{5}{\degree} (orange dots), \SI{10}{\degree} (blue dots), \SI{20}{\degree} (purple dots), \SI{40}{\degree} (black dots), and full \SI{360}{\degree} FF-HEDM patterns (green dots). The REI error bars show the range of REI values computed from 20 repeats. For clarity, the REI values are intentionally shifted by -0.1, 0, 0.1, 0.2, and 0.3 for $\Delta\omega$ of \SI{360}{\degree}, \SI{5}{\degree}, \SI{10}{\degree}, \SI{20}{\degree}, and, \SI{40}{\degree}, respectively. \tabref{tab:time_elapsed_rei_partial_data} shows the REI calculation time for each $\Delta\omega$ case. 

\begin{table}[tbhp]
    \centering
    \begin{tabular}{c|c|c}
        $\Delta\omega$ [deg] & Time for patch extraction [s] & Time for REI calculation [s] \\
        \hline
        \hline
        5 & 1 & 0.31 \\
        10 & 2 & 0.33 \\
        20 & 4 & 0.35 \\
        40 & 8 & 0.41 \\
        360 (full) & 72 & 1.2\\
    \end{tabular}
    \caption{REI calculation time for each $\Delta\omega$ case. Patch extraction time is computed assuming \SI{0.25}{\degree} \ome{} steps when acquiring FF-HEDM patterns and 20 frames per second processing time on the computing infrastructure used for this work.}
    \label{tab:time_elapsed_rei_partial_data}
\end{table}

In all four $\Delta\omega$ cases where a partial data set is used, the overall trends in the rare event indicator are consistent with those observed when the full data set is used to compute REI. Smaller $\Delta\omega$ results in larger REI error bars, most likely due to limited sampling and scatter in the FF-HEDM patterns. Nevertheless, the REI computed from $\Delta\omega$ of \degs{40} is very close to the benchmark REI. This implies that it is not necessary to acquire and analyze a full FF-HEDM data set acquired over \ome-range of \SI{360}{\degree} to compute a reliable REI. Furthermore, when combined with the REI calculation time (\tabref{tab:time_elapsed_rei_partial_data}), this observation shows that experimenters can consider a new mode of FF-HEDM data acquisition where the sample is continuously rotated and loading is not paused\footnote{The sample geometry and loading rate need to be moderated to a level that matches the diffraction pattern acquisition and rare event indicator determination rates.} and diffraction patterns are streamed without pausing\footnote{Such experimental modality will also require a stage stack that allows continuous \ome{} rotation such as the RAMS load frame or one that employs a slip ring system.}. Given the quick turnaround time for REI calculation, continuous quasi-static loading or even cyclic loading can be possible. As diffraction patterns are streamed, REI can also be streamed concurrently to provide a scalar actionable quantity to the experimenters to steer the course of the \insitu{} experiment. This is a loading modality that cannot be realized easily with the conventional metric such as FWHM or lattice strain (\figref{fig:steel_rei}) that requires a full reconstruction or, at the minimum, diffraction peak fitting. 

\subsection{REI with continuous loading}
\label{subsec:rei_with_cont_loading}
As alluded in \S\ref{subsubsec:_partial_data_ss_uq}, the ability to compute reliable REI from a partial FF-HEDM data set allows a new loading modality where loading is not paused during an \insitu{} FF-HEDM. We deploy our rare event detection framework on a previously published FF-HEDM data set that employed this loading modality \cite{pagan.2016ug} to quantify the crystallographic slip strength and study microcrack initiation and propagation in a Ti-7Al alloy. The experimental setup and sample geometry are similar to those used for the Ti64 data presented in \S\ref{subsec:ti_rei}. Another key difference to note in this FF-HEDM data set is that the \ome{} scan was broken up into six \SI{60}{\degree}-segments so as to test both the continuous loading modality and partial data REI. 

\begin{figure}[bth]
    \centering
    \includegraphics[width=\columnwidth]{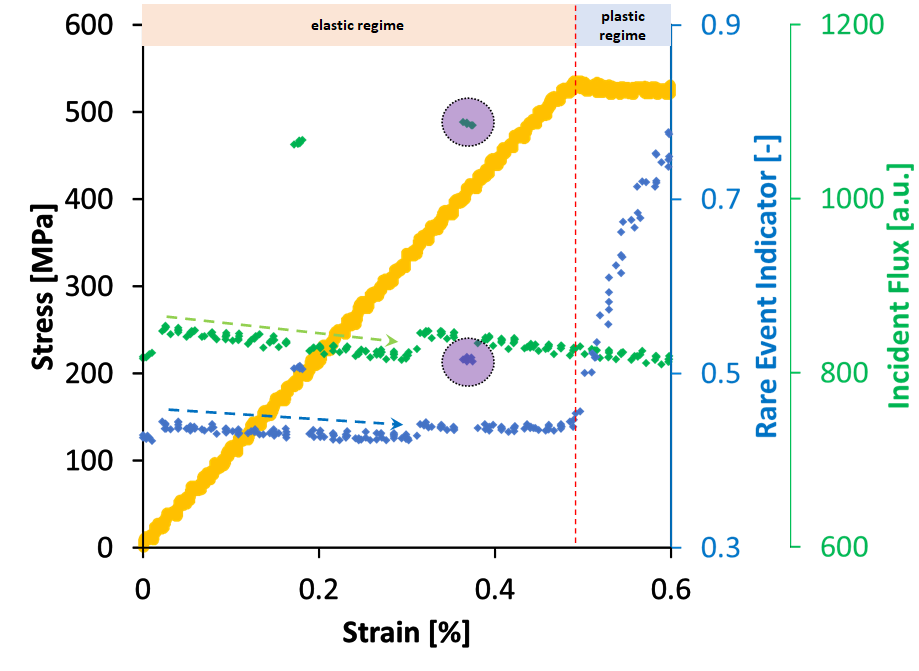}
    \caption{Ti7 alloy is continuously deformed in uniaxial tension while acquiring FF-HEDM patterns. REI shows significant changes in three different ways. It increases significantly when the material undergoes elastic-plastic transition. It also changes when the incident flux changes significantly, an example of such an instance is highlighted by the purple ellipses. Furthermore, REI is also sensitive to a gradual decay in the incident X-ray flux, an example of such an instance is highlighted by the dotted green (incident flux) and blue lines (REI).}
    \label{fig:Ti7_rei}
\end{figure}

\figref{fig:Ti7_rei} shows the macroscopic stress-strain curve, REI values computed over the course of the \insitu{} FF-HEDM, and incident x-ray flux. As anticipated, the rare event detection framework is capable of capturing the elastic-plastic transition from a partial FF-HEDM data set as indicated by the large increase in REI when the material transitions into the plastic regime. Furthermore, our rare event detection framework is capable of detecting changes in the incident x-ray flux. For instance, the REI values increase significantly with a large increase in the incident flux measured by an ion chamber installed in the x-ray beam path immediately before the sample (purple ellipse in \figref{fig:Ti7_rei}). We also observed a decay in REI consistent with the decay in the incident beam flux associated with the 324 singlets (Non Top-Up) APS storage ring operation modality when this data set was acquired. This trend is highlighted by the blue and green dotted arrow lines for REI and incident flux, respectively. 

This application example shows not only that our rare event detection framework can track changes to the diffraction peaks induced by crystallographic slip but also it is capable of detecting small changes to the incident beam flux. This is particularly useful for newer FF-HEDM modes that require much longer scan time (on the order of several hours) such as stitching HEDM \cite{Johnson2023}\footnote{Here, multiple field-of-view scans across a sample cross-section are stitched to interrogate a sample cross-section larger than the beam size available at the endstation} or point focus HEDM \cite{Li2023}\footnote{Here, multiple scans across a sample cross section are conducted with a 2D focused beam to acquire intra-granular microstructure and micromechanical state information.}. REI can be an independent metric that can indicate changes to the FF-HEDM patterns due to incident beam characteristic changes or material state changes (such as creep or relaxation) during these scanning techniques.

\section{Outlook}
We have demonstrated that our rare event detection framework is capable of detecting changes to the FF-HEDM diffraction peaks due to material state change and instrument changes such as incident flux and beam size. This framework is based on the unsupervised image representation learning and clustering algorithms. The resulting rare event indicator can be a metric that experimenters can use to make informed decisions about the course of their \insitu{}~FF-HEDM experiment instead of solely relying on a conventional stress-strain curve. Continuous loading combined with \insitu{}~FF-HEDM can be more accessible as experimenters do not have to rely on full reconstructions to decide on when to stop the loading and deploy higher resolution techniques. 

The rare event detection framework presented in this work can inform the experimenter \emph{when} a possible rare event is occurring. If the rare event detection framework can detect \emph{where} in the microstructure anomaly is occurring, it will be an additional piece of information that experimenters can use to steer their experiment. There are several avenues to identifying the location. For instance, it can be accomplished by creating a subset of patches from a particular set of \emph{grains of interest} and only monitoring the REI from those grains. 

It is also noteworthy that there are several petabytes of FF-HEDM data from more than a decade of user operation at the APS 1-ID beamline covering a wide range of alloy systems and multi-axial loading paths. Our intention is to use these data sets to further evaluate our rare event detection framework and report on the findings in the near future. 

\section{Methods}
\label{sec:methods}
We describe our rare event detection workflow,  hyperparameter tuning methods, and high-energy diffraction microscopy experiments.

\subsection{Rare event detection workflow}
\label{subsec:event_detection}

For several years, the field of event detection, also known as outlier detection or anomaly detection, has been a vibrant and dynamic area of study across diverse academic communities. Recently, the application of deep learning has opened new avenues for event detection, emerging as a critical and promising direction in this field \cite{deep_anomaly_survey}. A typical deep learning-based event detection involves a three-phase process. First, in the data preprocessing phase, the data is cleaned, normalized, and prepared for modeling. Second, during model training, a deep learning model, such as an autoencoder, is trained on a ``normal''  or ``baseline'' dataset to learn typical data patterns and accomplish the feature extraction task. Finally, in the event detection phase, this trained model is used to evaluate new data, identifying ``events'' by measuring deviations from the learned patterns.

\begin{figure}[bth]
    \centering
    \includegraphics[width=0.9\columnwidth]{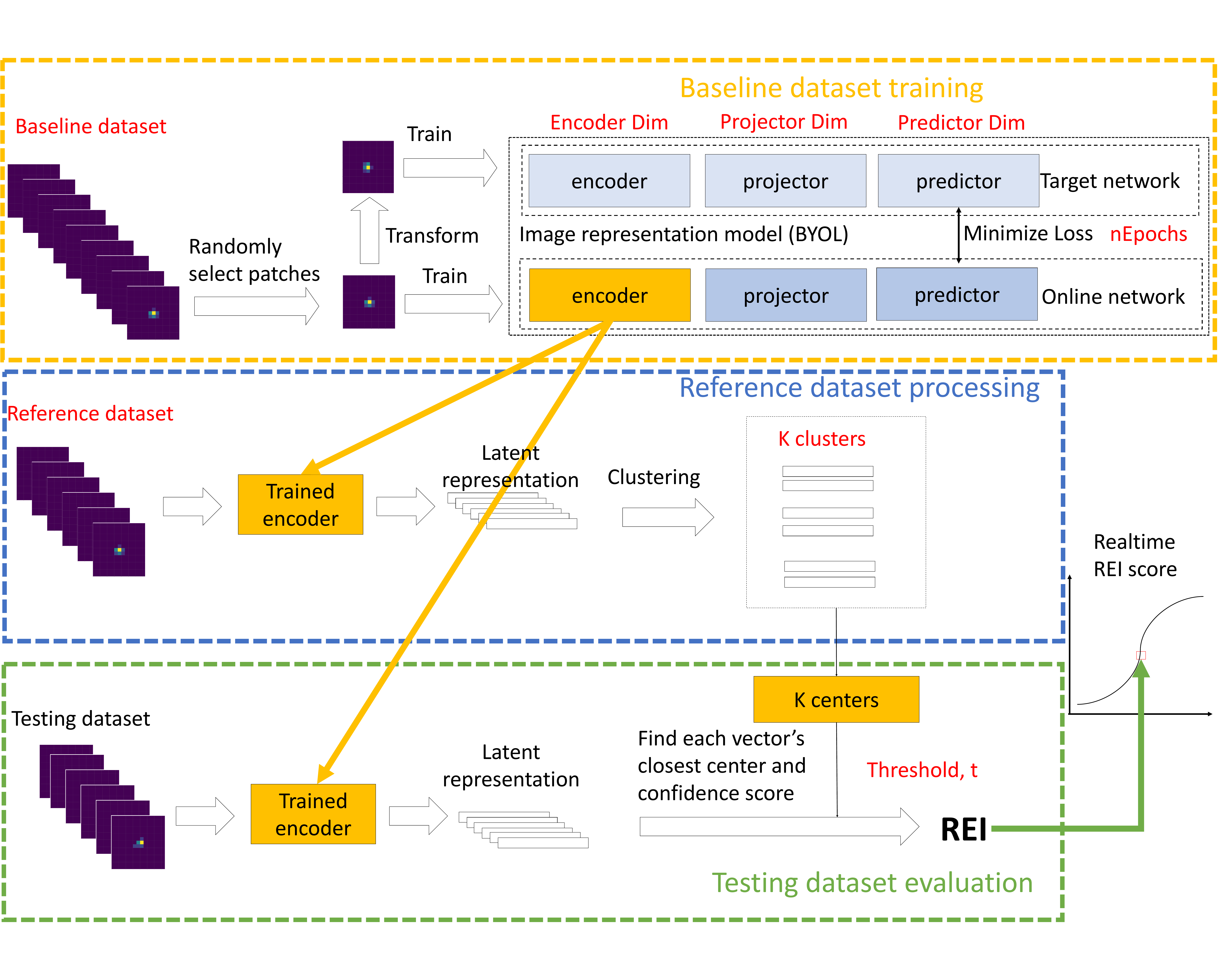}
    \caption{Rare event detection workflow with three phases. The first phase (orange rectangle), trains an image representation model (Trained encoder) using a baseline dataset for feature extraction. The trained encoder from the first phase is applied to a reference dataset followed by  K-means clustering algorithm to obtain K centers to characterize the reference dataset in the second phase (blue rectangle). The output of the trained encoder from the first phase, and the clustering model from the second phase, applied on the testing dataset, is thresholded to determine REI for the testing dataset in the third phase (green rectangle). The different hyperparameters at each step are shown in red text.}
    \label{fig:overview}
\end{figure}

In this study, as shown in Figure \ref{fig:overview}, we employ the BraggNN model \cite{BraggNN-IUCrJ} as our neural network model for feature extraction, specifically fine-tuned for image representation. To facilitate the unsupervised training of our novel image representation model, we adopt the BYOL (Bootstrap Your Own Latent) approach \cite{BYOL}. Additionally, we utilize the K-means clustering algorithm to compute the actionable information as a rare event indicator (REI). By separating the feature extraction tool and the event detector into two independent modules, our workflow can easily expand its support for event detection in various applications. Furthermore, our current design demonstrates promising results, showcasing the effectiveness of our event detection workflow on X-ray diffraction data.

We provide in the following a comprehensive overview of the different building blocks comprising our proposed workflow. \S\ref{subsubsec:event_detection_pre} elucidates the preprocessing steps involved in handling the HEDM diffraction images. \S\ref{subsubsec:event_detection_emb} describes the specifics of training the image representation model. We illustrate the modules associated with processing the reference dataset and evaluating subsequent experimental datasets in \S\ref{subsubsec:event_detection_reference} and \S\ref{subsubsec:event_detection_testing}, respectively.

\subsubsection{Data pre-processing}
\label{subsubsec:event_detection_pre}
In the context of our workflow, a fundamental objective of data preprocessing is to transform each dark-subtracted diffraction image extracted from a binary file into a set of peak patches that represent individual peaks. To accomplish this, we utilize the component analysis library provided by OpenCV \cite{opencv_lib}. A \textit{dataset} in our workflow is defined as the collection of peak patches (ranging in number of thousands to millions) extracted from diffraction images collected during a contiguous rotation segment (up to 360$^\circ$). Depending on the material state during acquisition, datasets are labeled \textit{baseline}, \textit{reference}, and \textit{testing} datasets. The \textit{baseline} and \textit{reference} datasets are typically any datasets collected before the start of the application of a macroscopic stimulus to the material. The \textit{testing} dataset is fed to the workflow as the material undergoes microstructural change.

\subsubsection{Image representation model training}
\label{subsubsec:event_detection_emb}

During the training phase of the image representation model, we utilize the Bootstrap Your Own Latent (BYOL) method \cite{BYOL} on the preprocessed baseline dataset. BYOL has emerged as a prominent self-supervised learning technique in deep learning, addressing the challenge of training neural networks without explicit labels or annotations. In \figref{fig:overview}, depicted by the orange rectangle, BYOL leverages two sets of neural networks: online and target networks. Each network comprises an encoder and a projector. For the encoder, we refine the BraggNN model by removing its fully connected layers. The projector, a two-layer fully connected neural network, maps the encoder's output to a lower-dimensional space, enhancing the model's generalizability.

The encoder's role is to extract the features of each data patch, while the projector aims to reduce the dimensionality of the encoder's output, facilitating improved model performance. During training, the online network learns to represent each peak patch in a latent space, while the target network maintains a moving average of the online network's parameters. The trained encoder is saved for use in subsequent phases. The core idea of BYOL is to encourage the online network to generate latent representations similar to those produced by the target network. By comparing these representations, BYOL effectively learns powerful representations for the given task. This approach has demonstrated remarkable success across various domains, enabling models to achieve state-of-the-art performance without the need for costly annotated data.

Even though our workflow only uses the trained encoder of the online network from BYOL, due to the structure and training of BYOL, the dimensions of the encoder, projector, and predictor networks, the number of epochs for training are important hyperparameters, which can impact REI when applied to the training dataset later on. \S\ref{subsec:hyperparameters} describes the hyperparameters in more detail.

\begin{figure}
    \includegraphics[width=\textwidth]{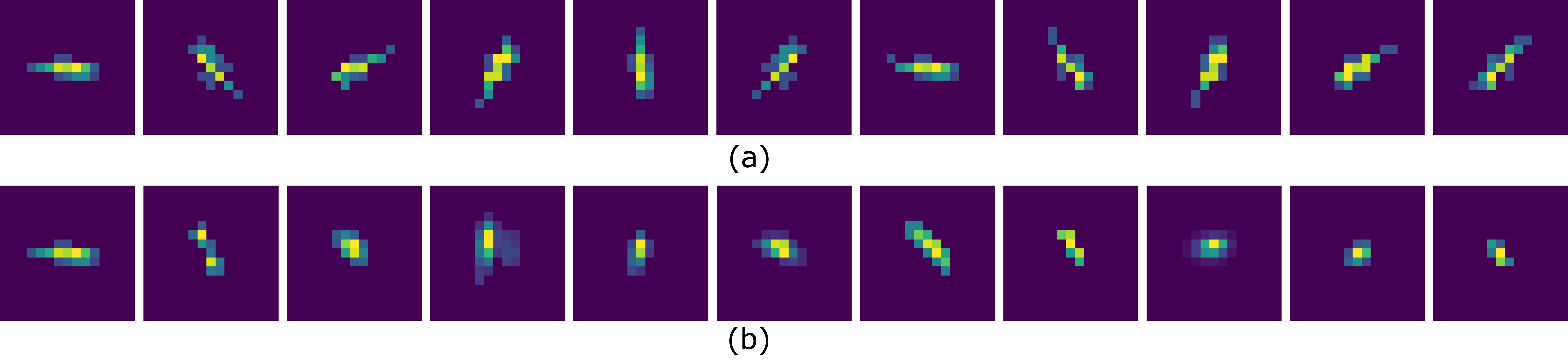}
    \caption{For a randomly picked peak (left-most), the median cosine distance between the peak and 10 transformed views (a) is 0.0058, while the median cosine distance with 10 randomly picked peaks (b) is 0.051, nearly 10x further. }
    \label{fig:rand-vs-aug}
\end{figure}

In each epoch of training, as illustrated in Figure \ref{fig:overview} with the orange rectangle, we randomly select a peak patch from the preprocessed baseline dataset. We then apply random transformations to the chosen patch. This improves the robustness of the model by reducing its sensitivity to such transformations, an example is shown in Figure \ref{fig:rand-vs-aug}. The median cosine distance\footnote{Median cosine distance is used as a metric to indicate similarity between two patches.} between a randomly selected peak and its transformed views (Figure \ref{fig:rand-vs-aug}a) is 0.0058, nearly 10x better than the distance (0.051) between the same peak and randomly selected different peaks (Figure \ref{fig:rand-vs-aug}b). The intuition underlying BYOL is to minimize the distance between the output of the target network and the online network, ideally converging to zero.

\subsubsection{Reference dataset processing}
\label{subsubsec:event_detection_reference}
In the reference dataset processing phase, our first step involves feeding all patches from the reference dataset into the trained representation model (encoder) and converting them into a collection of representation vectors. These vectors capture the essential features and characteristics of each patch.

Subsequently, we utilize the K-means clustering algorithm to group these representation vectors into K centers. This clustering process allows us to identify common patterns and clusters within the reference dataset. The resulting K centers (another hyperparameter) serve as representative points that summarize the distribution and variations present in the reference dataset.

The output of the reference dataset processing phase, comprising the K centers obtained through K-means clustering, is saved for the subsequent testing dataset evaluation phase. These centers provide a reference point for comparison and evaluation when analyzing the similarity and discrepancy between the testing datasets and the reference dataset.

\subsubsection{Testing dataset evaluation}
\label{subsubsec:event_detection_testing}
During the testing dataset evaluation phase, we utilize the trained representation model (encoder) to generate a collection of representation vectors for the testing dataset. Each vector captures the essential characteristics of a particular data instance (peak patch) within the testing dataset.

Next, we compute the Euclidean distances between each representation vector and the K centers obtained from the previous phase. This distance calculation allows us to determine the closest center for each vector, enabling us to assign it to a specific cluster. Additionally, we update the vector's closest center to obtain an assignment to K centers and its corresponding confidence level.

To quantify the uncertainty in the assignments, we introduce the REI score. This score is determined by calculating the percentage of representation vectors whose confidence level is below a predefined threshold value, denoted as $t$ (another hyperparameter). A lower confidence level indicates higher uncertainty in the assignment to a specific center. Accordingly, the REI score values will fall in between 0 and 1.

By repeating this procedure for all testing datasets, we obtain rare event indicators for each dataset. These scores provide an indication of the level of uncertainty associated with the assignment of data instances to the K centers. Collectively, these rare event indicator values form a figure that summarizes the uncertainty across all datasets, allowing for a comprehensive assessment of the performance and reliability of the event detection workflow.

\subsection{Hyperparameter tuning}
\label{subsec:hyperparameters}

Hyperparameters play a pivotal role in the machine learning workflow, determining the configuration and behavior of a model.
In our REI workflow, the hyperparameters during model training are as follows (numbers in parentheses are optimized values used here): dataset used for training the encoder, dataset used for training the clustering model, encoder dimension (32), projector dimension (64), predictor dimension (64), number of epochs for training the encoder, confidence threshold, \textit{t}, and number of clusters, \textit{K}.

As shown in \S\ref{subsec:rei_robustness_study}, REI is more sensitive to changes in beam size, beam flux, and rotation steps than to changes in position and starting rotation angles. Thus, the datasets used for training the encoder (\textit{baseline datasets}) and for training the clustering model (\textit{reference datasets}) only included datasets with fixed beam size, beam flux, and rotation steps.

To determine the optimal number of training epochs for our model, we employ a confidence summation approach. The idea behind this method is straightforward: a well-trained image representation model should be able to identify pairs of transformed patches within a set of random patches.

Our confidence summation process unfolds as follows: At the conclusion of each training epoch, we randomly select 100 patches from our dataset, convert them into 100 representation vectors using our model, and calculate a confidence score for each one. This score is derived from the distances between a specific patch representation and 10 other patches. To compute the confidence score, we use the formula:

\begin{equation}
    \text{Confidence} = \frac{|D_1 - D_2|}{D_2}, \\
\end{equation}
where $D_1$ and $D_2$ are the shortest and second shortest Euclidean distances, respectively, among all pairs of representation vectors. If the distance between two transformed patches is not the closest one among these 10 patches, we assign a confidence score of -1. This process enables us to evaluate how well patches differ from their nearest neighbors in terms of distance. Finally, we sum up all 100 confidence scores for all representations of patches. 

\begin{figure}[bth]
    \centering
    \includegraphics[width=0.7\columnwidth]{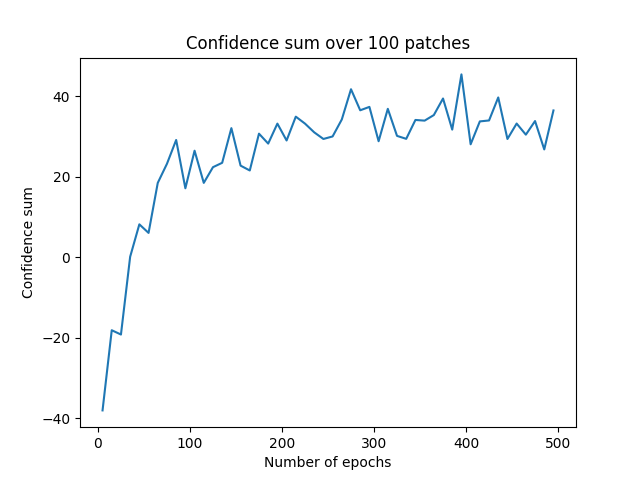}
    \caption{Accuracy plot of the image representation model.}
    \label{fig:epochs_tune}
\end{figure}

In our experiments using the stainless steel dataset described in Section \ref{subsec:ss_uq}, we observe that the confidence score stabilizes after approximately 100 epochs, as depicted in Figure \ref{fig:epochs_tune}. Therefore, we opt to set the number of training epochs to 100 for our image representation model.

\begin{figure}[bth]
    \centering
    \includegraphics[width=0.7\columnwidth]{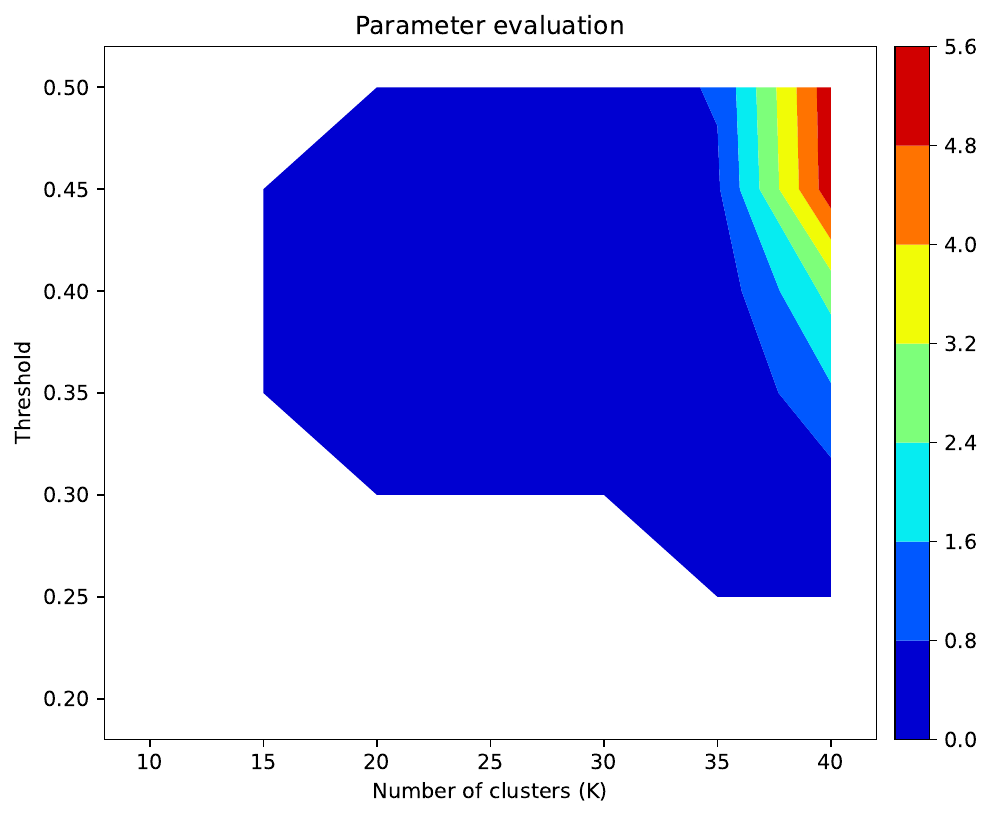}
    \caption{Contour plot of REI sensitivity as a function of number of clusters $K$ and threshold $t$.}
    \label{fig:param_tune}
\end{figure}

To determine the optimum number of clusters, $K$, and threshold, $t$, we will describe hyperparameter tuning as it was applied to the stainless steel dataset described in $\S$\ref{subsec:ss_uq}. The performance of the model is evaluated by computing the change in REI when the material first undergoes plastic deformation (two data points in the purple ellipse in Figure \ref{fig:steel_rei}). REI sensitivity, \emph{REI}\textsubscript{sensitivity} (shown in Figure \ref{fig:param_tune}) is maximized during hyperparameter tuning, defined as follows:

\begin{equation}
    REI_{\textrm{sensitivity}} = \dfrac{REI^{\textrm{min}}_{\textrm{plastic}} - REI^{\textrm{max}}_{\textrm{elastic}}}{\Delta REI_{\textrm{plastic}}},
\end{equation}
where $REI^{\textrm{min}}_{\textrm{plastic}}$ is the smallest REI value for the data point after the onset of plastic deformation and $REI^{\textrm{max}}_{\textrm{elastic}}$ is the largest REI value for the data point before the onset of plastic deformation in the elastic regime and $\Delta REI_{\textrm{plastic}}$ is the spread in REI (due to local microstructure variations) among different volumes scanned for the data point after the onset of plastic deformation.

Figure \ref{fig:param_tune} shows a contour plot of \emph{REI}\textsubscript{sensitivity} as a function of number of clusters and confidence threshold. Only positive values of \emph{REI}\textsubscript{sensitivity} are shown.\footnote{Negative \emph{REI}\textsubscript{sensitivity} means the model fails to detect the onset of plastic deformation at that data point.} It can be seen that a higher number of clusters and higher confidence threshold results in higher \emph{REI}\textsubscript{sensitivity}. As depicted in Figure \ref{fig:param_tune}, the selection of K as 40 and the threshold as 0.5 lead to the highest sensitivity of \emph{REI}\textsubscript{sensitivity}, reaching 5.6. 

\subsection{High-energy diffraction microscopy experiments}
\label{subsec:hedm}
\begin{figure}[bthp]
    \centering
    \includegraphics[width=\columnwidth,trim=3mm 0 0 0,clip]{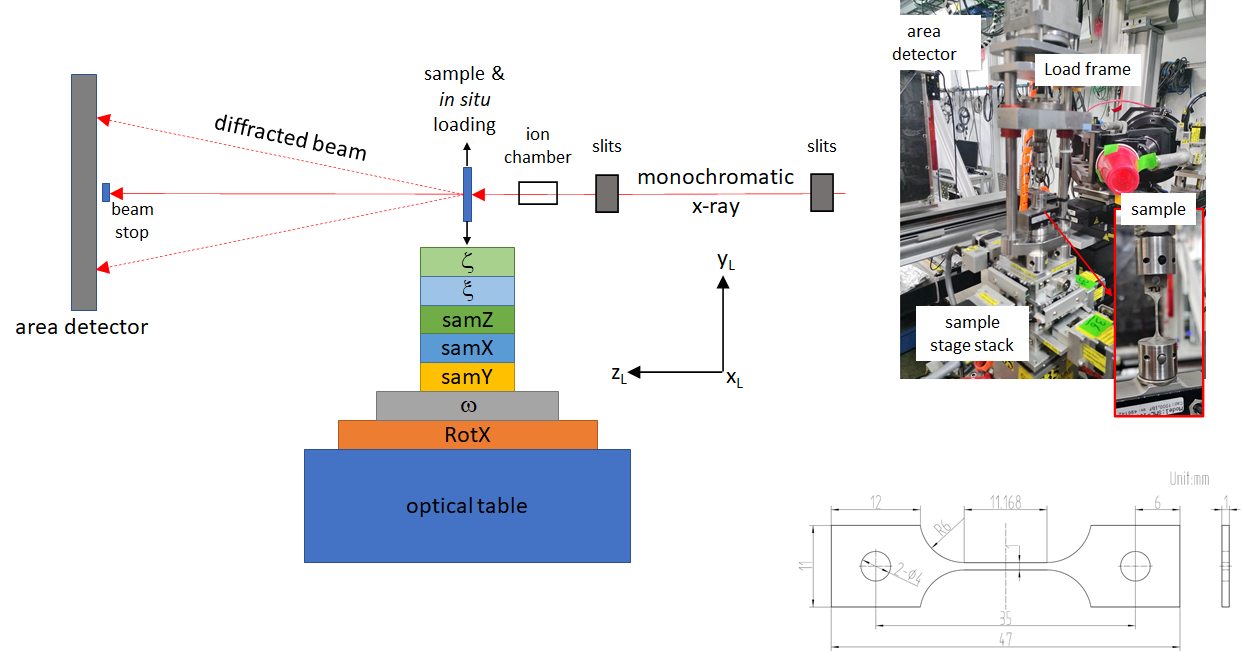}
    \caption{A schematic of the FF-HEDM instrument geometry. The stage stack consists of translation and rotation capabilities to align the \ome-rotation axis and the sample to the x-ray beam. The picture shows the setup and sample geometry used to acquire the \insitu{} FF-HEDM data for the 304L-SS sample.}
    \label{fig:hedm_exp_schematic}
\end{figure}

The FF-HEDM data sets presented here were acquired at the 1-ID beamline of the Advanced Photon Source (APS) at Argonne National Laboratory. The FF-HEDM instrument geometry (\figref{fig:hedm_exp_schematic}) is detailed in \cite{Park2021}. Its main features are:
\begin{itemize}
    \item Monochromatic, high-energy, synchrotron X-rays were used as the probe to interrogate the material.
    \item Transmission geometry with an area detector \cite{Lee.2007us} was used to capture the intensity of the diffraction spots in reciprocal space. \tabref{tab:hedm_exp_parameters} lists the x-ray energies and sample-to-detector distances used.
    \item The incident X-ray beam had a rectangular shape. The beam size along \xl{} was sufficiently large to illuminate the cross-section of the sample in its gauge section and the beam size along \yl{} was varied depending on the experiment's main objectives and required resolution in between layers; it was varied only when investigating the effect of beam size on the rare event indicator. 
    \item \figref{fig:hedm_exp_schematic} shows an example sample geometry used for the uniaxial tension testing. When possible, a set of \SI{30}{\micro\meter} cube gold markers \cite{shade.2016pf} were attached on the sample gauge section surface so that identical volume of material was interrogated throughout the \insitu{} HEDM experiment.
    \item The sample was rotated about \yl{} at a constant angular speed over an \ome{} range of \SI{360}{\degree} while a set of diffraction patterns were acquired. Typically, this set consisted of 1440 frames, each covering an angle of \SI{0.25}{\degree}. It took approximately \SI{6}{\min} to acquire a set. 
    \item In a typical \insitu{} FF-HEDM experiment at the APS, Microstructural Imaging using Diffraction Analysis Software (MIDAS) \cite{Sharma2012ASetup,Sharma2012AGrains,MIDAS_www} is used to analyze the set of diffraction patterns and extract a 3D map. At moderate deformation levels, it takes approximately \SI{8}{\min} to acquire the 3D map from a data set on a high performance computing cluster. 
\end{itemize}

\begin{table}[tbhp]
    \centering
    \begin{tabular}{c|c|c}
        & X-ray energy (keV) & Sample to detector distance (mm) \\
        \hline
        304L-SS (\S\ref{subsec:ss_uq}) & 71.68 & 803 \\
        Ti-6-4 (\S\ref{subsec:ti_rei}) & 71.68 & 1163 \\
        sand (\S\ref{subsec:sand_uq}) & 71.68 & 1601 \\
        Ti-7 (\S\ref{subsec:rei_with_cont_loading}) & 61.33 & 756 \\
    \end{tabular}
    \caption{FF-HEDM instrument parameters.}
    \label{tab:hedm_exp_parameters}
\end{table}

\section{Acknowledgements}
\label{sec:ack}
This research used resources of the Advanced Photon Source, a U.S.\ Department of Energy (DOE) Office of Science user facility at Argonne National Laboratory and is based on research supported by the DOE Office of Science-Basic Energy Sciences, under Contract No. DE-AC02-06CH11357. This work was also supported by the DOE Office of Science, Office of Basic Energy Sciences Data, Artificial Intelligence and Machine Learning at DOE Scientific User Facilities program under Award Number 08735 (``Actionable Information from Sensor to Data Center"). We express our thanks to Professor Aaron Stebner of Georgia Institute of Technology for sharing the 304L-SS material, Dr.~Paul Shade and his team at the U.S.\ Air Force Research Laboratory for fabricating the gold markers for the 304L-SS sample, Professor Matthew Kasemer and his team at the University of Alabama for sharing the Ti64 data, and Professor Ryan Hurley and his team at Johns Hopkins University for sharing the sand data.

\section{Code and data availability}
The data used here is archived on the APS Data Management System and is available on request. The code is available on request.

\section{Author contributions}
AA, HS, and ZL conceived the machine learning methods. AA, ZL, and WZ developed the machine-learning software with guidance from IF and RK. JSP, PK, and HS performed the synchrotron experiments for 304L-SS data set with assistance from AM and ZL. WZ performed the data analysis under the guidance of JSP, PK, and HS. All the authors discussed the results and wrote the manuscript.

\bibliographystyle{ieeetr}
\bibliography{main}

\end{document}